\documentclass[letterpaper, 10 pt, conference]{ieeeconf}  

\IEEEoverridecommandlockouts                              

\usepackage{graphics} 
\usepackage{graphicx}
\usepackage{epsfig} 
\usepackage{amsmath} 
\usepackage{amssymb}  
\usepackage{todonotes}
\usepackage{subfig}
\usepackage{epsfig}

\usepackage{todonotes}
\usepackage{balance}

\graphicspath{{./}{./figures/}}

\title{\LARGE \bf
Learning Human-Aware Path Planning with Fully Convolutional Networks*
}

\author{No\'{e} P\'{e}rez-Higueras$^{1}$, Fernando Caballero$^{1}$ and Luis Merino$^{1}$
\thanks{*This work is partially supported by the EC-FP7 under grant agreement no. 611153 (TERESA) and the project PAIS-MultiRobot funded by the Junta de Andaluc\'{i}a (TIC-7390)}
\thanks{$^{1}$No\'{e} P\'{e}rez-Higueras, Fernando Caballero and Luis Merino are with School of Engineering, Universidad Pablo de Olavide, Crta. Utrera km 1, Seville, Spain
        {\tt\small noeperez@upo.es, fcaballero@us.es, lmercab@upo.es}}%
}

\begin{document}

\maketitle
\thispagestyle{empty}
\pagestyle{empty}

\begin{abstract}
This work presents an approach to learn path planning for robot social navigation by demonstration. We make use of Fully Convolutional Neural Networks (FCNs) to learn from expert's path demonstrations a map that marks a feasible path to the goal as a classification problem. The use of FCNs allows us to overcome the problem of manually designing/identifying the cost-map and relevant features for the task of robot navigation. The method makes use of optimal Rapidly-exploring Random Tree planner (RRT$^{*}$) to overcome eventual errors in the path prediction; the FCNs prediction is used as cost-map and also to partially bias the sampling of the configuration space, leading the planner to behave similarly to the learned expert behavior. The approach is evaluated in experiments with real trajectories and compared with Inverse Reinforcement Learning algorithms that use RRT$^{*}$ as underlying planner.
\end{abstract}

\section{Introduction}\label{sec:intro}

Nowadays, mobile robots are becoming part of our daily lives. Robots must be prepared for sharing the space with humans in their operational environments. Therefore, the comfort of the people and the human social conventions must be respected by the robot when is navigating in the scenario. Besides, the robots have to be coherent and maintain the legibility of their actions \cite{kruse_ras13}. This is called human-aware navigation.  

First approaches for including human-awareness into robot navigation were based on hard-coding some specially-designed constraints in the motion planners in order to modify the robot behavior in the presence of people \cite{KirbySF09, DBLP:journals/trob/SisbotMAS07}. However, the task of "social" navigation is hard to define mathematically but easier to demonstrate. Thus, a learning approach from data seems to be more appropriate \cite{luber_iros12}.

In the last years, several contributions have been presented regarding the application of learning from demonstrations to the problem of human-aware navigation \cite{argali09learning,luber_iros12}. One successful technique to do that is Inverse Reinforcement Learning (IRL) \cite{Ng:2000}: the observations of an expert demonstrating the task are used to recover the reward (or cost) function the demonstrator was attempting to maximize (minimize). Then, the reward can be used to obtain a similar robot policy. 

Several IRL approaches for the robot navigation task can be found in the literature. In \cite{Vasquez2014} a experimental comparison of different IRL approaches is presented. Also in \cite{roman14}, IRL is applied to transfer some human navigation characteristics into a robot local planner. A similar work is done in \cite{Kim2016}, where the densities and velocities of pedestrians observed via RGB-D sensors are used as features in a reward function learned from demonstrations for local robot path planning.   

Most of the IRL approaches frame the problem as a Markov Decision Process (MDP) where the goal is to learn the reward function of the MDP. The use of MDPs presents some drawbacks, as the difficulty of solving the MDP at each learning step in large state spaces. This limits the applicability of most of these approaches to small states spaces. Some authors have tackled these MDP limitations from different points of view. A graph-based approximation is employed in \cite{OkalIcra16} to represent the underlying MDP in a socially normative robot navigation behavior task. Another example is \cite{kuderer2015learning} where IRL is used to learn different driving styles for autonomous vehicles using time-continuous splines for trajectory representation. In \cite{kretzschmar_ijrr16} the cooperative navigation behavior of humans is learned in terms of mixture distributions using Hamiltonian Markov chain Monte Carlo sampling. Other authors have tried to replace the MDP by a motion planner. In \cite{perez-higueras2016_icsr_learning}, an approach based on Maximum Entropy IRL \cite{ziebart08IRL} is employed to learn the cost function of a RRT$^*$ path planner. Also in \cite{ShiarlisICRA17} an adaptation of the Maximum Margin Planning approach \cite{Ratliff_MMP_2006} to work with RRT$^*$ planner is presented.  

Moreover, the IRL techniques present other problems when the task to be learned is complex, like the human-aware navigation task. In these cases we have to manually design the structure of the reward and features involved in the task. In many cases, the designed reward probably is not able to recover the underlying demonstrated behavior properly. Even though the weights of the cost function can be learned from expert demonstrations, we still need to define a set of features functions that describe the space states and the relation between them. While for small problems this is doable, for many realistic and more complex problems, like human-aware navigation, it is hard to determine.

However, in the last years, the use of deep neural networks in IRL problems is bringing good results. The neural networks, as a function approximators, permit the representation of nonlinear reward structures in domains with high dimensionality. Some authors have already applied deep networks to the problem of human-aware robot navigation. In \cite{Chen2017}, a Reinforcement Learning approach is applied to develop a socially-aware collision avoidance system where a deep network is employed to learn multi-agent crossing trajectories. Also, Deep Q-networks have been used in \cite{Sharifzadeh2016}, to solve the MDP step of an IRL algorithm for the task of driving a car. Furthermore, the well-known Maximum Entropy IRL \cite{ziebart08IRL} algorithm is extended in \cite{DBLP:journals/corr/WulfmeierOP15} to use deep networks, and its application to path planning in urban environments is presented in \cite{WulfmeierIROS2016}.    
  

In this work we consider a different approach. We propose a learning from demonstration strategy for the task of human-aware path planning that avoids the explicit representation and definition of the cost function and its associated features. The proposed method does not follow the classical IRL approach. Instead, the problem is formulated as a classification task where a Fully Convolutional Network (FCN) is used to learn to plan a path to the goal in the local area of the robot in a supervised way, using the demonstrations as labels. This general approach in then combined with an optimal path planner to solve efficiently the task and to ensure collision-free paths. Thus, the contribution of this paper is twofold: 1) a novel approach for learning human-aware path planning from demonstrations based on Fully Convolutional Networks and 2) the combination of this information with an RRT$^*$ planner to enhance the planner capabilities while it behaves similarly to the expert behavior.    

The paper is structured as follows. 
Section \ref{sec:learning} describes the methodology followed for learning the task of human-aware path planning proposed. Then, Section \ref{sec:netvalidation} shows the results of experiments for validation of the deep learning approach, and Section \ref{sec:eval} presents a set of experiments including a comparison with other learning algorithms in realistic situations. Finally, Section \ref{sec:conclusions} summarizes the paper contribution and outlooks future work.

\section{Learning to Plan Paths with FCNs} \label{sec:learning}

In this section, we present our approach to learn from demonstrations to plan human-aware paths in 2D environments for robot social navigation. We propose using a Fully Convolutional Network to learn, from the expert's path demonstrations, to predict the correct path to the goal according to the actual obstacles and people in the vicinity of the robot, without explicitly defining the environment features used. Unlike Deep IRL approaches, that try to learn the cost function that the expert is following in a MDP framework, our goal is to directly predict the path to the goal given the information from obstacles and persons.

The path predicted by the FCN is then combined with a RRT$^{*}$ planner in order to perform the navigation task efficiently and to prevent the robot for collisions or prediction fails.

\subsection{Input and output of the network}\label{sec:net-input}

\begin{figure}[!tb]
\centering
\includegraphics[width=0.49\columnwidth]{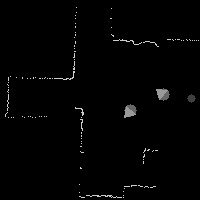}
\includegraphics[width=0.49\columnwidth]{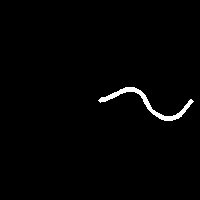}
\caption[Input to the network and label for learning]{Left. Example of input grid to the network. Obstacles, goal position and people positions and orientations are included. Right. Respective label used for learning with the path to the goal.}
\label{fig:input}
\end{figure}

We consider a local navigation task in a 2D space centered in the robot of size $10 \times 10~m^2$. Thus, the sensor data together with the detected people position and orientation is used as input to the network. No other information is employed, so the network has to derive the features of the task based on the provided state information. 

The sensor data based on laser scans and point-clouds is projected in a 2D grid of $200 \times 200~pixels$ with resolution of $0.05~m/pixel$. As we are considering the task of social navigation, we need to be able to detect people. So, a people detection system is employed to provide information about the position an approximate orientation in the scenario. This data is also included in the 2D grid, where the people are marked using circles and triangles to indicate the orientation. The goal is also marked on the grid as a small circle. This grid is used as a gray-scale image where the background color is black and each element previously described takes a different gray intensity. These values are normalized in order to be in the range $[0,1]$ before serving as input to the FCN. An example of the input gray-scale image of the network can be seen on the left image of the Fig. \ref{fig:input}. 


The output of the network is an approximate path from the center (robot position) to the goal marked on a grid, which has the same size as the input. We can see an example on the right image of the Fig. \ref{fig:input}.  

\begin{figure}[!tb]
\centering
\includegraphics[width=1.0\columnwidth]{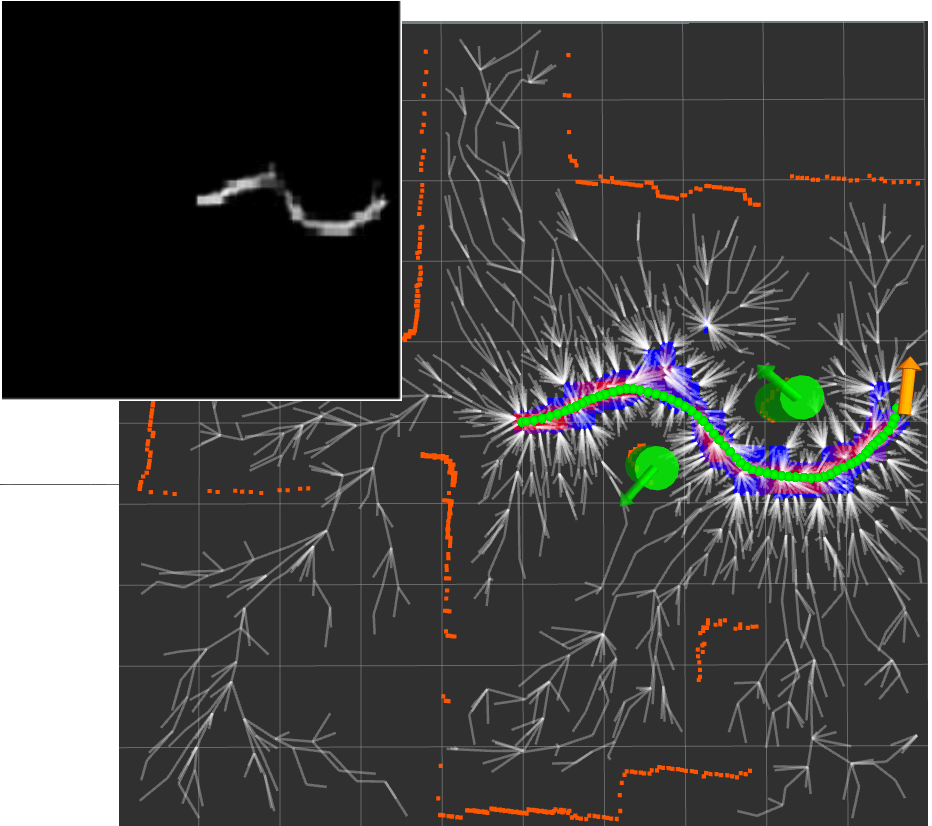}
\caption[Network output and respective RRT$^*$ path]{Upper left. Example of output plan of the network. Down right. RRT$^*$ path using the network prediction with a bias of the sampling of $70~\%$. Robot position is the center of the image. Orange lines shows the obstacles. People position and orientation are indicated with green cylinders and arrows. The goal is represented with the golden arrow. The planner tree is drawn in white color.}
\label{fig:nav}
\end{figure}

Finally, the path prediction from the network and the resulting RRT$^*$ path for the same example presented in Fig. \ref{fig:input} are shown in Fig. \ref{fig:nav}. As can be seen, the resulting path is very similar to the demonstrated one. \\       

\subsection{Network architecture}

\begin{figure}[!tb]
\centering
\includegraphics[width=1.0\columnwidth]{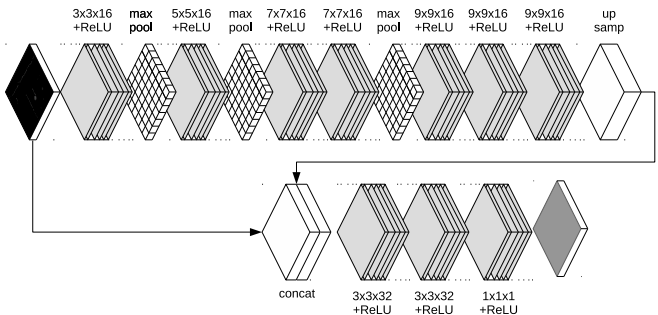}
\caption[Network architecture]{Proposed network architecture. The network is composed by two branches, a global-coarse branch that extract high level features of the input grid and a local-fine branch that make use of these features and the original input grid to generate the final path.}
\label{fig:net}
\end{figure}

Path planning problems are specially characterized by the existence of a goal destination in its formulation. This information is critical to optimize the path searching policies. The robot needs to know where to go prior to plan the best trajectory from its current position. 

The importance of the goal destination has been considered into the design of the proposed network architecture, presented in Fig. \ref{fig:net}. The architecture is inspired by the global-coarse to local-fine deep learning architecture presented in \cite{Eigen:2014}. Thus, the proposed structure is divided into two major branches: a global-coarse estimation that sequentially subsample the input grid while applying larger kernels in order to extract global high level features of the input, and the local-fine branch that make use of the extracted global features and the original input grid to build the final path considering local information and global constraints.  



Notice how the proposed network does not make use of fully connected layers, which significantly reduces the total number of parameters to one hundred thousand approximately. Instead, an output layer with 1x1 kernel size is used to generate the output grid with the path. 

\subsection{Integration with the RRT* planner}

The final final step consist in combining the path predicted by the FCN with a RRT$^{*}$ planner in order to perform the navigation task efficiently and to prevent the robot for collisions or prediction fails. 

The planner uses the path prediction in two ways: first, as a cost function to connect the tree nodes; and second, to partially bias the sampling of the state space, taking a higher percentage of samples from the areas where the plan is. This allows to guide the sampling process efficiently to areas of interest and thus, reaching an optimal path faster. 

The percentage of samples drawn from the path prediction or from a uniform distribution is an important parameter. Keeping a percentage of uniform sampling allows the planner to be still able to find a path to the goal when the prediction is not complete or fails. In this work, we have used a $70~\%$ of samples from the path prediction.


\begin{figure*}[!tb]
  \centering
  \begin{tabular}[b]{cc|cc|cc}
    \includegraphics[width=.3\columnwidth]{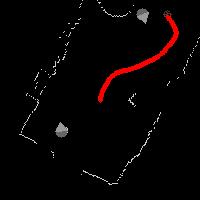} &
    \includegraphics[width=.3\columnwidth]{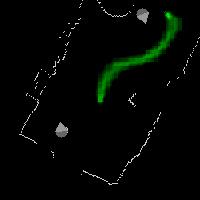} &
    \includegraphics[width=.3\columnwidth]{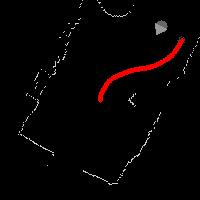} &
    \includegraphics[width=.3\columnwidth]{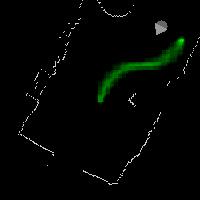} &
    \includegraphics[width=.3\columnwidth]{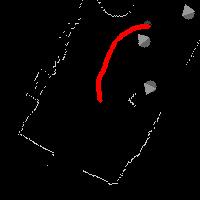} &
    \includegraphics[width=.3\columnwidth]{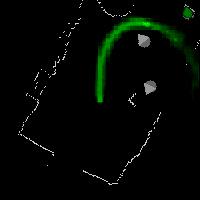} \\ \noalign{\smallskip}\hline\noalign{\smallskip}\noalign{\smallskip}
    \includegraphics[width=.3\columnwidth]{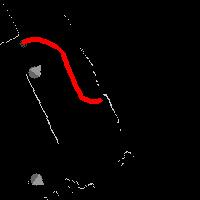} &
    \includegraphics[width=.3\columnwidth]{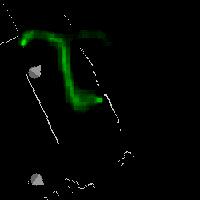} &
    \includegraphics[width=.3\columnwidth]{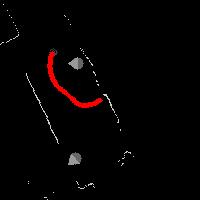} &
    \includegraphics[width=.3\columnwidth]{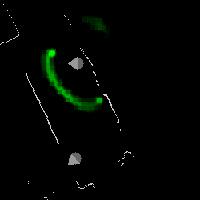} &
    \includegraphics[width=.3\columnwidth]{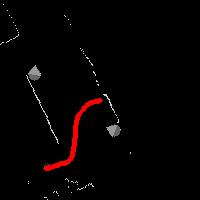} &
    \includegraphics[width=.3\columnwidth]{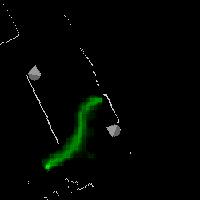} \\ 
  \end{tabular} 
  \caption{Visual comparison of $6$ pairs of labels and network predictions from the testing dataset. For each pair, the label is represented in red line on the left image, and the prediction is shown in green color on the right image.}
  \label{fig:netresults}
\end{figure*}



\section{Deep Network Validation} \label{sec:netvalidation}

We first validate the deep learning approach by testing the performance of the network prediction.
A dataset of $10500$ trajectories has been used to validate the capability of the proposed FCN to learn to plan paths. The dataset has been created by randomly generating positions and orientations for the robot, the goal and for the people around in different scenarios within a large map. 

Regarding the generation of demonstration paths to the goals in the scenarios, a RRT$^*$ planner with a pre-defined cost function has been employed. In particular, a weighted linear combination of five features is used as cost function similarly to the one employed in \cite{perez-higueras2016_icsr_learning} for robot social navigation. These features are based on distance metrics as the Euclidean distance to the goal and to the closest obstacle. Also metrics regarding the people in the vicinity of the robot are taken into account. The Euclidean distances and orientations with respect to the people in the scene are used through three Proxemics functions placed in front, back and sides of each person.

From the dataset, $10000$ trajectories are used in the learning process, while the remaining $500$ trajectories are reserved for testing the network after learning finishes. Also, during the learning, a number of $100$ trajectories were taken from the learning set as a validation set for overfitting checking. The results in terms of Mean Squared Error (MSE) for the different sets of trajectories, are presented in Table \ref{tab:netval}. As can be seen, the error committed in the testing set keeps low regarding the error reached in the learning set.   

\begin{table}[t]
\caption{Mean Squared Error (MSE) reached in the different stages with different set of trajectories}
\label{tab:netval}
\begin{center}
\begin{tabular}{llll}
\hline\noalign{\smallskip}
   &  Learning set  & Validation set & Testing set \\ 
 \noalign{\smallskip}\hline\noalign{\smallskip}
MSE & $0.0073$ & $0.0067$ & $0.0094$ \\ 
 \hline\noalign{\smallskip}
\end{tabular}
\end{center}
\end{table}

Furthermore, a visual comparison of some network predictions corresponding to some of the $500$ trajectories of the testing set are presented in Fig. \ref{fig:netresults}. The images are presented in pairs with the demonstrated path in red in the left image, and the respective prediction on the right image in green. As can be observed, the predictions fit the expert's paths very well.  

\begin{figure}[!tb]
  \centering
  \begin{tabular}[b]{cc}
    \includegraphics[width=.40\columnwidth]{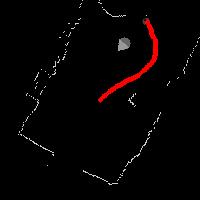} &
    \includegraphics[width=.40\columnwidth]{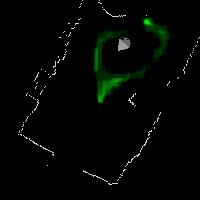} \\ \noalign{\smallskip}\hline\noalign{\smallskip}\noalign{\smallskip}
    \includegraphics[width=.40\columnwidth]{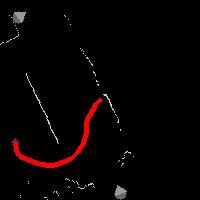} &
   \includegraphics[width=.40\columnwidth]{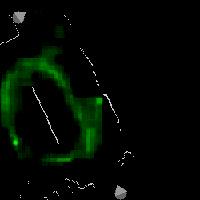}  
  \end{tabular} 
  \caption{Example of two cases where the network learn to plan in different valid homotopies. The demonstrated path is shown in red color on the left images while the respective prediction is shown in green color on the right side images.}
  \label{fig:homotopies}
\end{figure}

An interesting and unexpected outcome of the proposed network is its capability to generate more than one valid path when the goal can be similarly reached by following different homotopies, even when the learning has been made with a single trajectory per example. Figure \ref{fig:homotopies} shows two examples of prediction in two valid homotopies. This information is easily handled by the RRT$^*$ planner which will select the shorter path. 

The implementation code of the network and all the datasets used in the experiments can be found in the Github repository of the UPO Service Robotics Lab, in the module \textit{upo\_fcn\_learning}\footnote{https://github.com/robotics-upo/upo\_fcn\_learning}.

\section{Path Planning Evaluation} \label{sec:eval}

The feasibility of the proposed approach is demonstrated for human-aware planning by learning in a small dataset of real robot trajectories. Then we compare the resulting trajectories with a ground-truth set and two IRL algorithms of the state of the art that learn the cost function of a RRT$^{*}$ planner as a weighted linear combination of features.

\subsection{Dataset}

A dataset of $300$ trajectories has been recorded in different controlled scenarios with static people around and where the robot was remotely controlled by a human expert who tried to perform a correct human-aware navigation to the goal.

This dataset is sorted in $3$ sets, where $250$ demonstration paths are used for learning and $50$ are used for testing the results. The difference between sets is that the $50$ paths chosen for testing (and thus the 250 for demonstration) are different for each set.

 
Is noteworthy that, in terms of deep learning with FCNs, a dataset of $250$ demonstrations for learning human-aware navigation can be considered very small. We will show that even with this small information the presented approach is able to equal or overcome the result of two state-or-the-art IRL algorithms.

\subsection{State-of-the-art algorithms}

The performance of our approach is tested against two IRL algorithms of the state of the art: RTIRL \cite{perez-higueras2016_icsr_learning} and RLT \cite{ShiarlisICRA17}. Both try to learn the weights of a linear combination of features used as cost function of a RRT$^*$ planner using the same set of demonstrated trajectories. The first one is based on a Maximum Entropy approach \cite{ziebart08IRL} that replace the MDP by an RRT$^*$ planner; while the second one is an adaptation of the Maximum Margin Planning algorithm (MMP) \cite{Ratliff_MMP_2006} to work with a RRT$^*$ planner.

We have used with these algorithms the extended set of features specifically designed for robot social navigation in \cite{perez-higueras2016_icsr_learning}. The set is a compound by five features. Three of them are based on distances and orientations between the robot and the people in the scene and are coded through three Gaussian functions placed in front, back and side of each person in the scene. Then, two more features measuring the distance to the goal and distance to the closest obstacle are also considered. It is important to remark that our approach, on the contrary, is fed with just the raw laser data and the position and orientation of people in the form of an image, as explained in Section \ref{sec:net-input}.

The implementation of the IRL algorithms used for comparison can be found in the module \textit{upo\_nav\_irl}\footnote{https://github.com/robotics-upo/upo\_nav\_irl} of the Github repository from the UPO Service Robotics Lab.

\subsection{Metrics}   

In order to compare the planned paths with the ground-truth paths, we use a metric based on a directed distance measure between two paths:

\begin{equation} 
D(\zeta_1, \zeta_2) = \frac{ \sum_{i=1}^N d(\zeta_1(i), \zeta_2) }{N} 
\end{equation} \label{eq:metric}

\noindent where the function $d(\zeta_1(i), \zeta_2)$ calculates the Euclidean distance between the point $i$ of path $\zeta_1$ and its closest point on the path $\zeta_2$, and $N$ stands for the number of points of the path $\zeta_1$.
This distance in then combined to obtain a final metric of path comparison $\mu$:

\begin{equation}
\mu(\zeta_1, \zeta_2) = \frac{ D(\zeta_1, \zeta_2) + D(\zeta_2, \zeta_1) }{2} 
\label{eq:avg_metric}
\end{equation}

Moreover, a comparison between the difference in the feature counts of the ground-truth paths and the planned paths is also employed, as the feature counts play a key role in the IRL approaches \cite{perez-higueras2016_icsr_learning,ShiarlisICRA17}. The feature count of a path is defined in Equation \ref{eq:featurecount}, where $f(\zeta(i))$ indicates the vector of features values for point $i$ of path $\zeta$, and $\Vert \zeta(i+1) - \zeta(i) \Vert$ is the Euclidean distance between the points $i$ and $i+1$ of the path $\zeta$. Even though our FCN approach does not make use of such features, we also compute their counts for the resultant trajectories.

\begin{equation} 
F(\zeta) = \frac{ \sum_{i=1}^{N-1} f(\zeta(i)) + f(\zeta(i+1))}{2} \Vert \zeta(i+1) - \zeta(i) \Vert
\end{equation} \label{eq:featurecount}

\subsection{Comparative results} 

We first train the RLT, RTIRL and FCN approaches with the learning set, and then compare the results using the testing set. We obtain the results for the three different data combinations described above.

\begin{figure}[!tb]
\centering
\includegraphics[width=1.1\columnwidth]{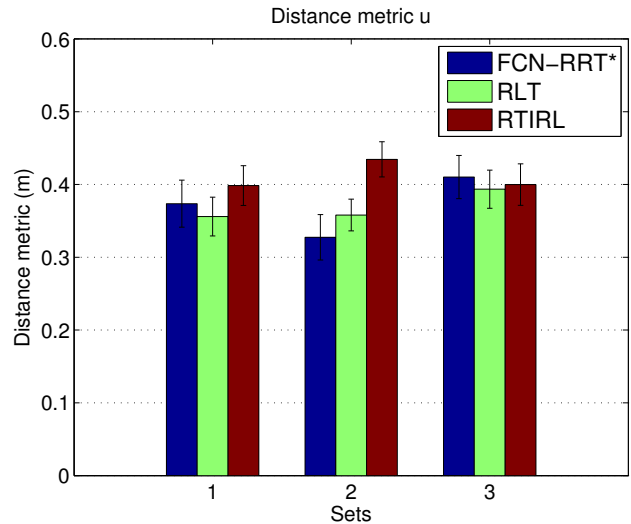}
\caption[Distance metric $\mu$]{Average distance metric $\mu$ and standard errors for the three approaches in the three sets of trajectories of test.}
\label{fig:metric}
\end{figure}

First, the average values of the distance metric (\ref{eq:avg_metric}) between the planned paths and the $50$ ground truth trajectories of each testing set are shown in Fig. \ref{fig:metric}. As can be seen, the performance of the three approaches is very similar. Even though no further information is provided to the FCN approach, it is able to learn an adequate representation of the task, and equals the results obtained by the other two algorithms that use a pre-defined set of features engineered for human-aware robot navigation.

\begin{figure*}[!tb]
\centering
\includegraphics[width=0.67\columnwidth]{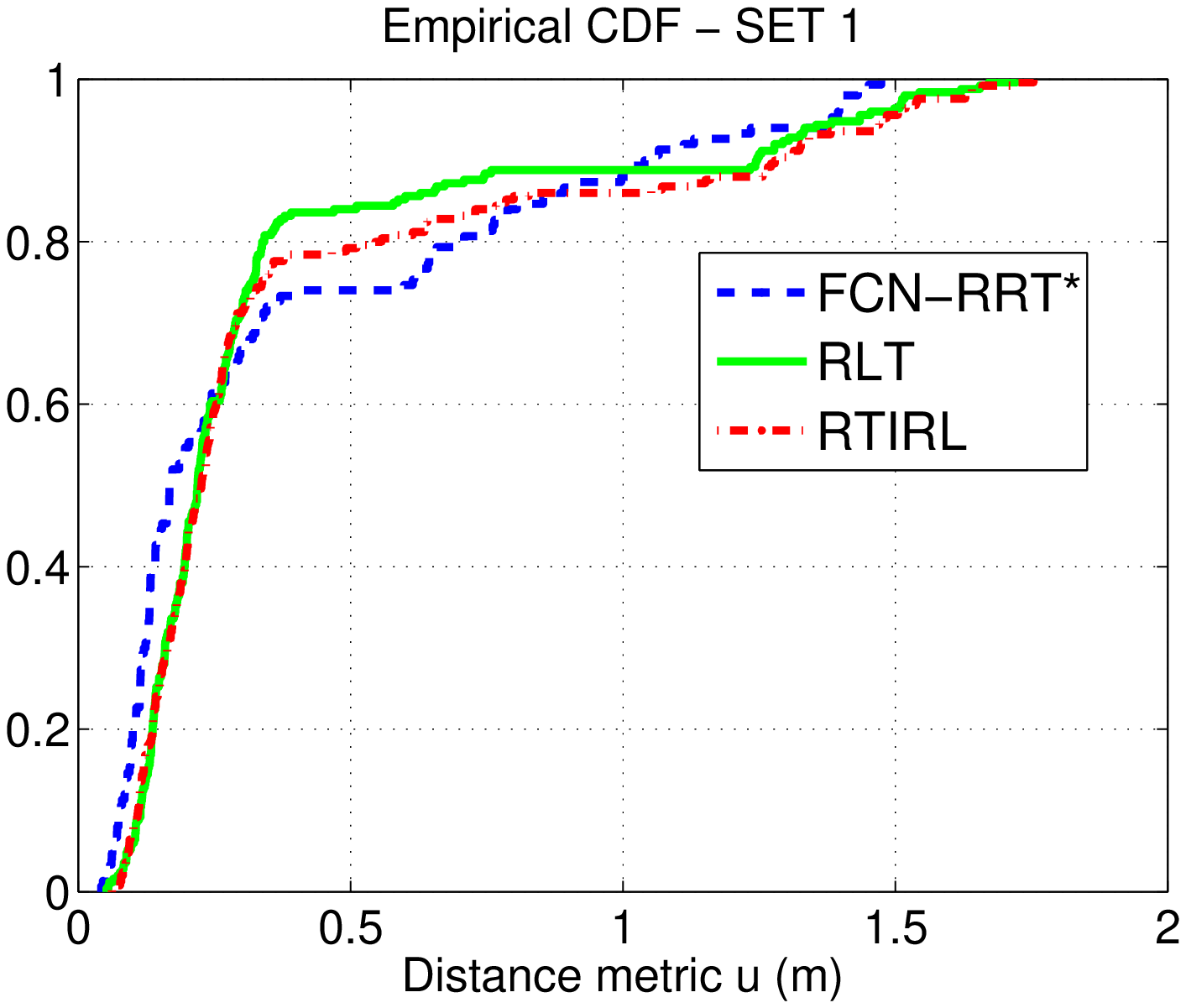}
\includegraphics[width=0.67\columnwidth]{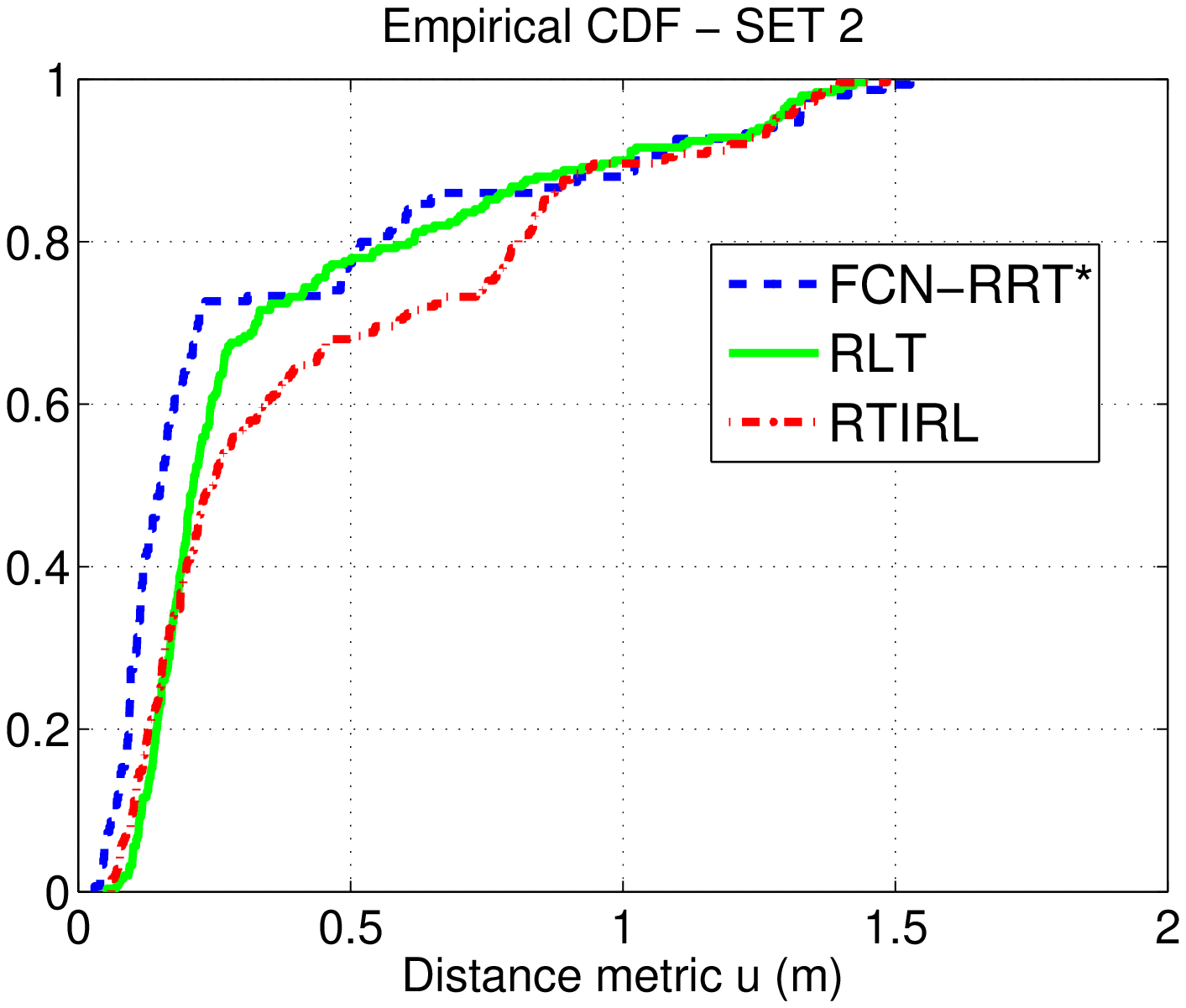}
\includegraphics[width=0.67\columnwidth]{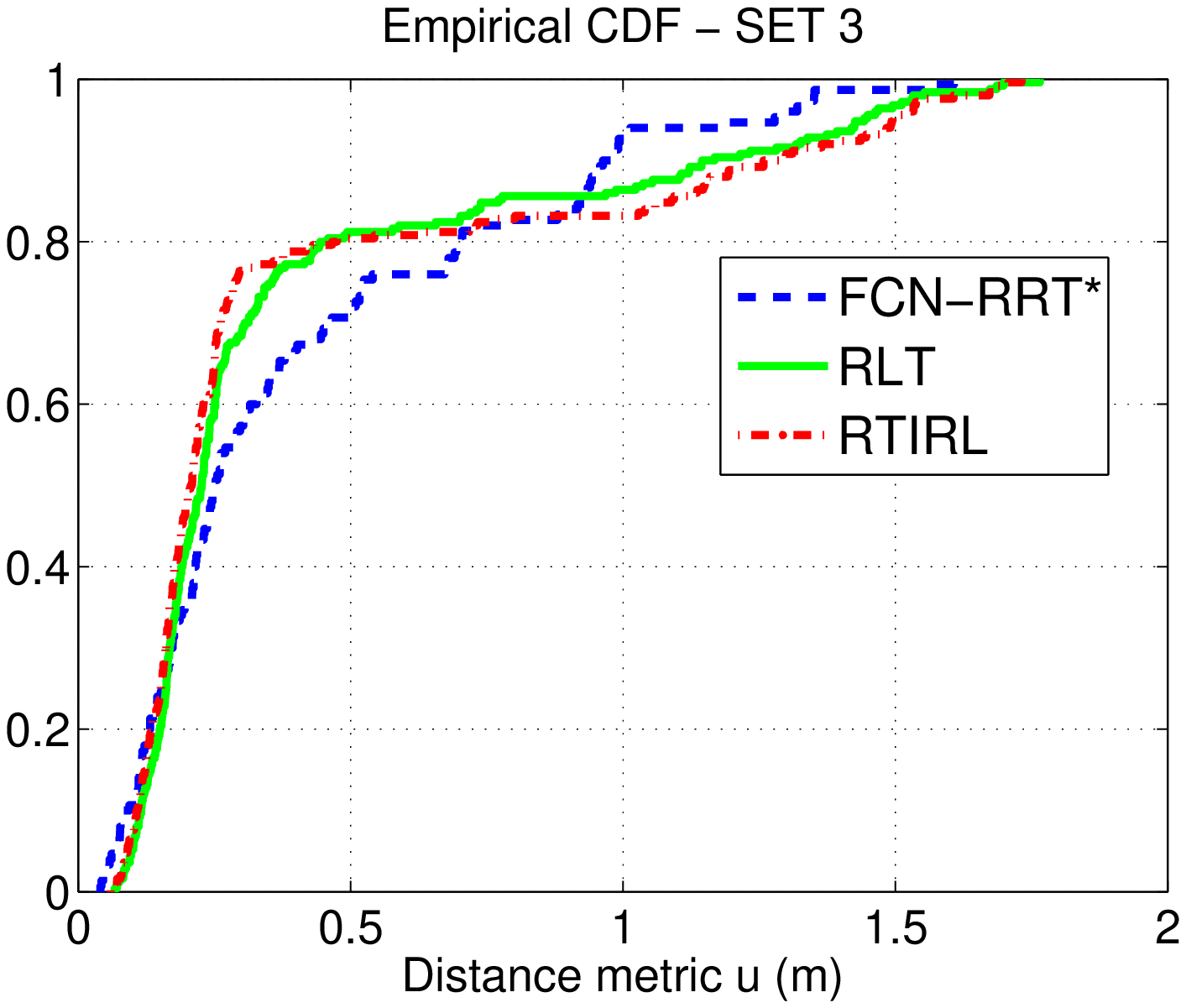}
\caption[Distance metric $\mu$]{Cumulative error in the distance metric $\mu$ for the different approaches in the three sets of trajectories for testing.}
\label{fig:CDF_mu}
\end{figure*}

Figure \ref{fig:CDF_mu} shows a comparison of the cumulative density functions of the distance metric $\mu$ for the different approaches in the three sets of trajectories for testing. Again, the performance of the different approaches is very similar to our approach.   

\begin{figure*}[!tb]
\centering
\includegraphics[width=0.67\columnwidth]{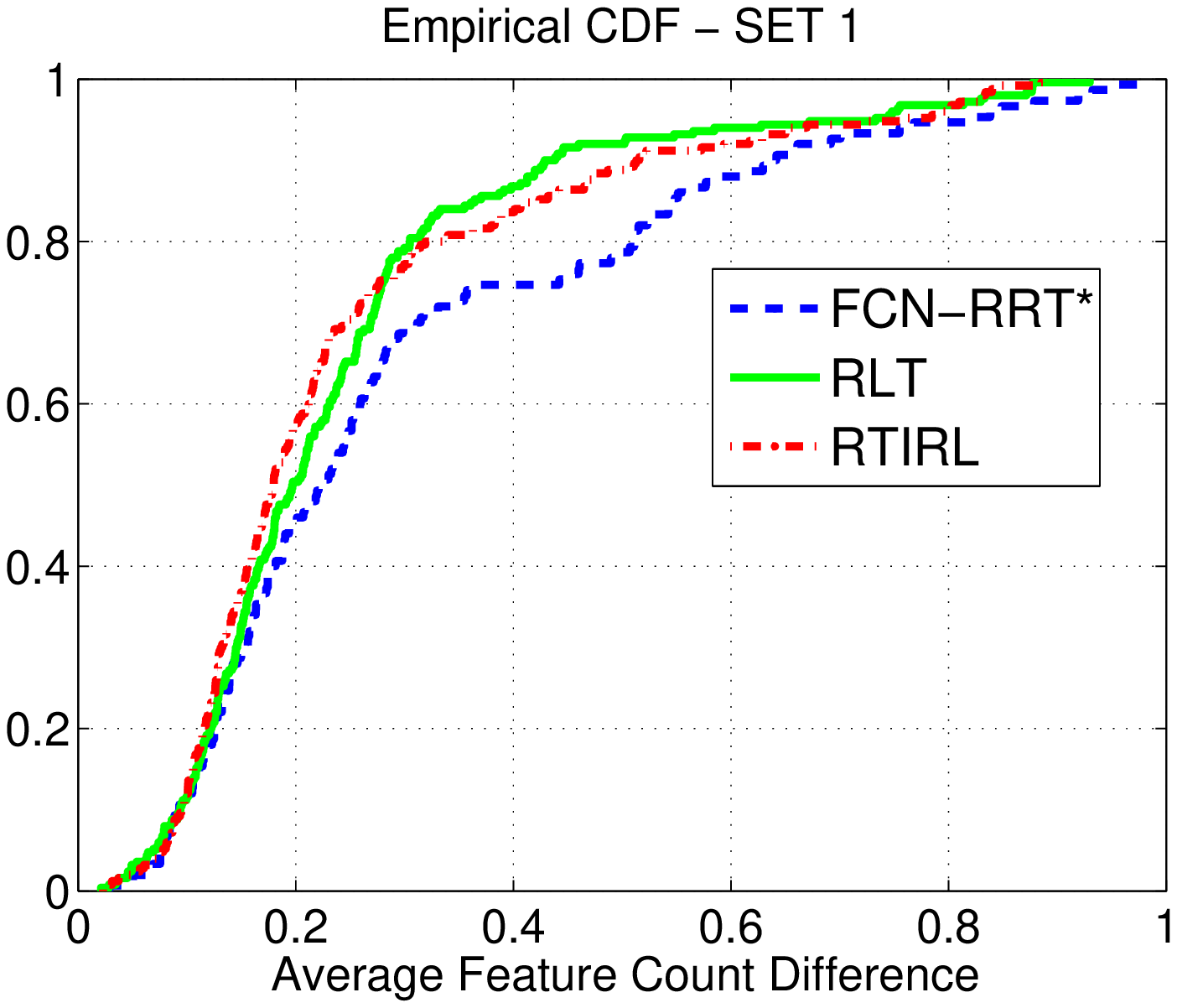}
\includegraphics[width=0.67\columnwidth]{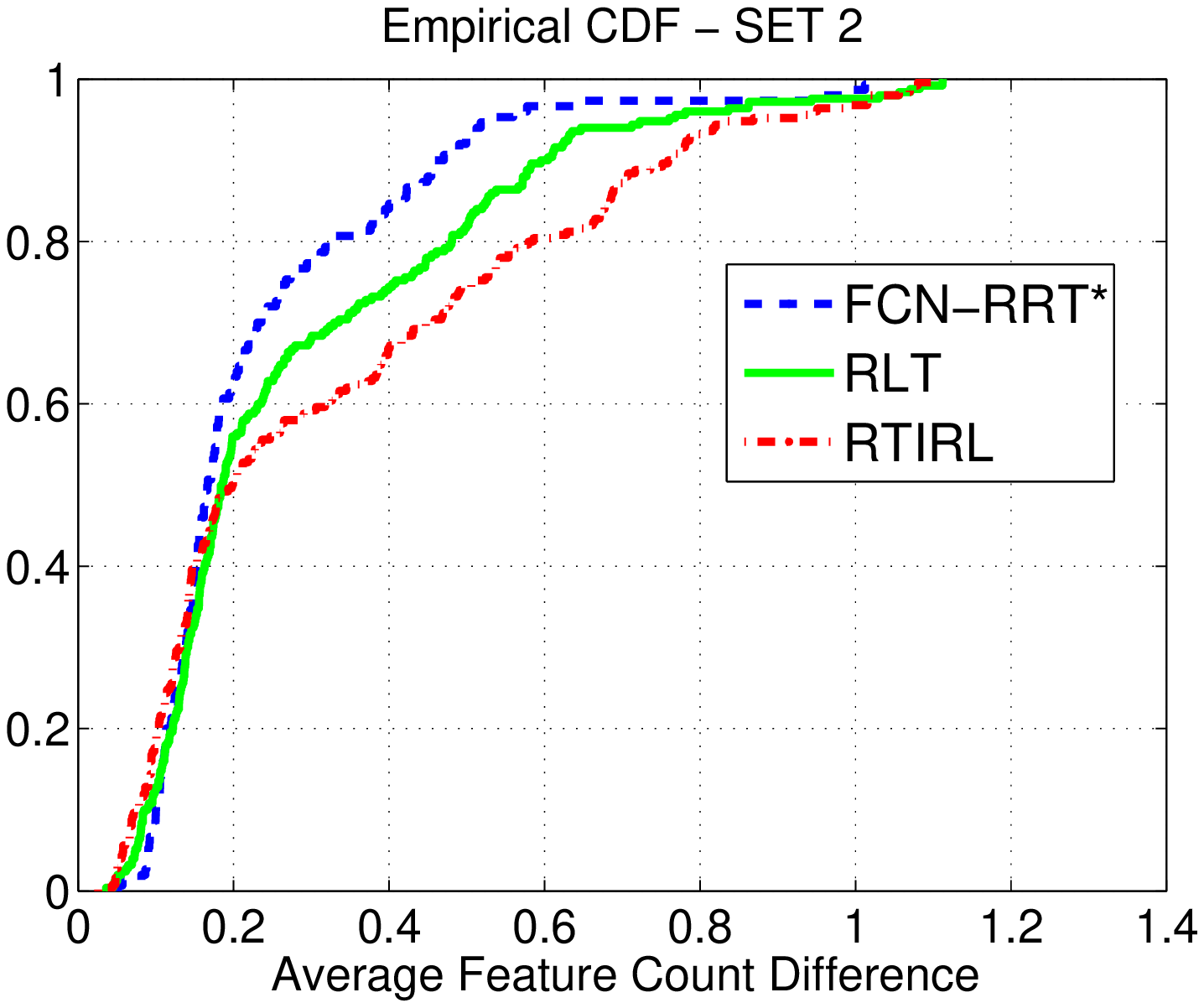}
\includegraphics[width=0.67\columnwidth]{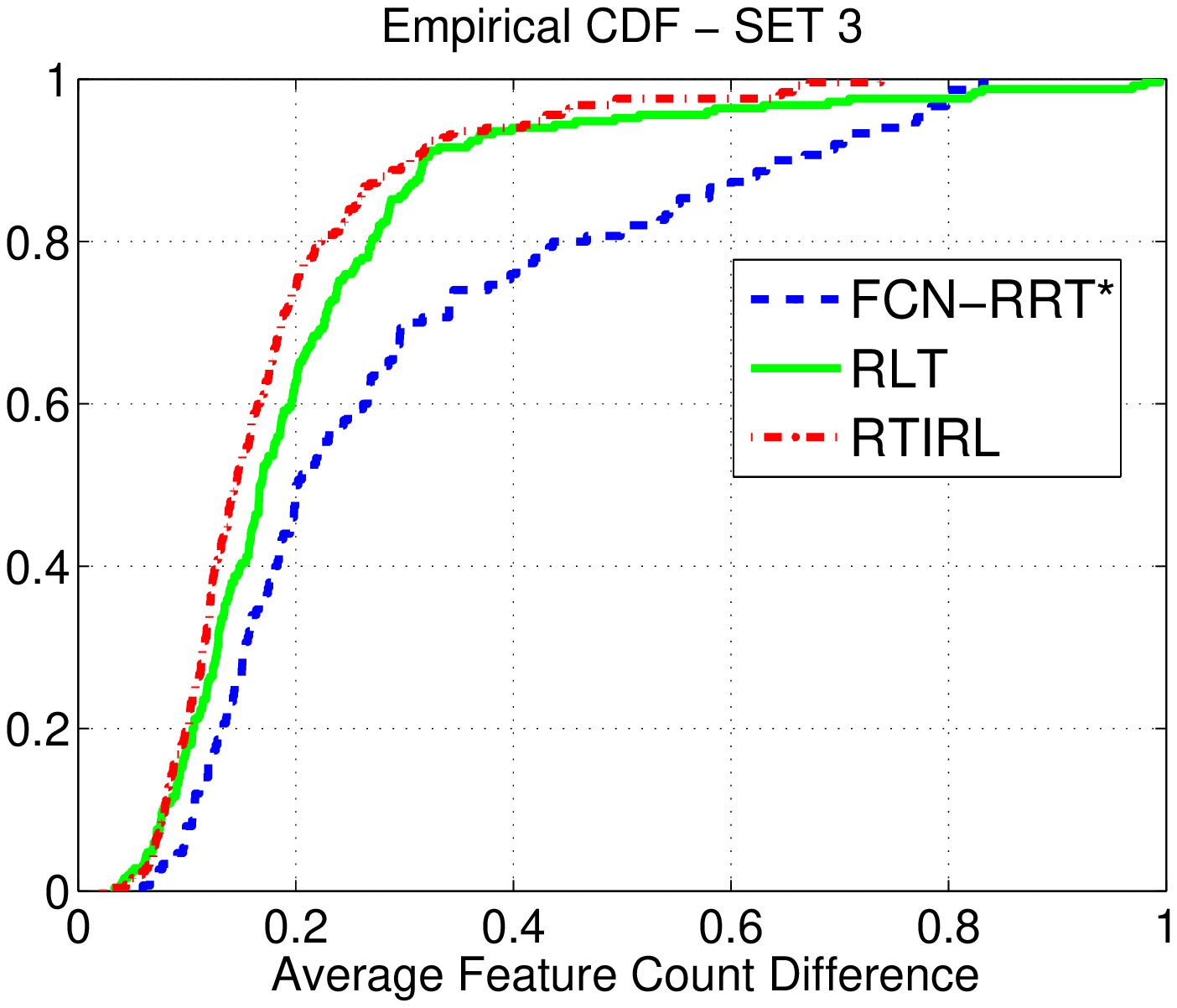}
\caption[Distance metric $\mu$]{Cumulative error in the average feature count difference for the $3$ approaches in the $3$ sets of trajectories for testing.}
\label{fig:CDF_fc_diff}
\end{figure*}

Finally, we make a comparison in terms of the manually-designed features. We compare the differences in the feature counts of the planned paths with respect to the ground-truth paths of the three testing sets. Figure \ref{fig:CDF_fc_diff} shows the cumulative density functions of the average feature count difference. It is interesting to see how the RTIRL and RLT obtain similar results while our method under scores in two of the sets, but the resulting paths are very similar according to Fig. \ref{fig:CDF_mu}. This is an expected result given that the navigation features have not been specified to our approach, so the FCN could likely converge to another set of features that also leads to good imitation of the demonstration trajectories.

\section{Conclusions and Future Work} \label{sec:conclusions}

This paper presented an approach for learning human-aware path planning based on the integration of FCNs and RRT$^*$. The introduction of FCNs to learn the path planning based on demonstration allows to address the problem without the need of hand-crafted social navigation features as many other IRL approaches in the state of the art. Additionally, the integration of the predicted path with RRT$^*$ guarantees an optimal feasible path no matter the situation.

The full approach has been tested with a set of real trajectories and benchmarked with state-of-the-art algorithms for human-aware navigation learning with good results. Our approach offers very similar metrics without defining the navigation features. In addition, the neural network prediction has been also successfully tested with a large dataset in order to validate the predicted paths.

Future work will consider comparing the results with different network architectures in order to benchmark our network with respect existing solutions. Also, learning in dynamic scenarios will be considered, so that the approach can exploit people trajectories to improve the quality of the path planning in such scenarios.









\balance
\bibliographystyle{IEEEtran}      
\bibliography{references}   

\end{document}